# Automatic Detection of Ringworm using Local Binary Pattern (LBP)

Srimanta Kundu
Jadavpur University, Kolkata-32
srimantacse@yahoo.co.in

Nibaran Das*
Jadavpur University, Kolkata-32
nibaran@cse.jdvu.ac.in
*corresponding author

Mita Nasipuri
Jadavpur University, Kolkata-32
mitanasipuri@gmail.com

**ABSTRACT**

In this paper we present a novel approach for automatic recognition of ring worm skin disease based on LBP (Local Binary Pattern) feature extracted from the affected skin images. The proposed method is evaluated by extensive experiments on the skin images collected from internet. The dataset is tested using three different classifiers i.e. Bayesian, MLP and SVM. Experimental results show that the proposed methodology efficiently discriminates between a ring worm skin and a normal skin. It is a low cost technique and does not require any special imaging devices.

## 1. INTRODUCTION

Computer aided detection/diagnosis (CAD) [1] of medical diseases is an active research field which includes analysis of digital images of affected regions. A successful CAD system provides useful information for the clinician's diagnostic support. But the development of proper CAD system is a challenging mission. Despite research efforts from different frontline research groups over the last century, it has remained an active research avenue due to the diversity as well as high complexity of the problems. Though some systems have been developed to address several diseases, but those are not sufficient enough. Most of the researches are going on to identify different types of cardiac diseases, cancer etc which have alarming effect on human existence. But pitifully very few mentioned have been found to address skin related problems. Elif Derya Ubeyli et al. worked on automatic detection of Erythro-squamas disease. They used k-mean clustering [2] and neuro-fuzzy inference [3]system for that purpose. Automatic detection of poultry skin worm was implemented by Du Z, Jeong et al. They used the concept of band selection of hyper spectral images [4] to implement that. Automatic cancer detection was implemented by H.B.kerle et al. using vector quantization for segmentation method. [5] Infrared hyper spectral reflectance imaging was used by Welling wang et al. for detection of sour skin disease in Vidalia sweet onions [6].

But still identification of proper skin disease is challenging enough due to symptom similarities with that of others. Among different skin diseases, Dermatophytosis or Ringworm, a clinical condition, is caused by fungal infection. The funguses form a ring-shaped rash outside the body and remain alive during the infection stage. There are several worms involved to develop a ring worm. In general, around twenty percent of the population is affected with ring worm at any given moment. The disease is very much common among sports persons and the persons living in a worm, humid climate having direct contact with active lesions on someone else or having weakened immunity system (due to having diabetes, leukemia, or AIDS). Misdiagnosis and improper treatment of it may lead to tinea incognito where funguses are spread out by far without any control. Therefore, it is very important to identify the ring worm at an early stage. But to the best of our knowledge, there is no CAD system yet developed for recognition of ringworm automatically. From this motivation, we have developed an automatic computer aided ring worm detection scheme based texture patterns of different region of skin. Hence, it is important to identify a set of discriminative features strong enough to represent the proper textures of the skins. In the present work we have introduced Local Binary Pattern (LBP) [7] based texture features for recognition of ringworm. The texture features of the skin are computed from the digital image of the skin using LBP method. On the basis of these features, decision is made whether the skin is affected with ringworm or not. We have evaluated the disease based on the acquired images where the presence of ring worm is positive or negative. For this purpose we have used three different types of classifiers namely Bayesian classifier, Multi Layer Perceptron (MLP) and Support Vector Machine (SVM). And final decision is taken using majority voting schemes.

The paper is organized as follows: Section 2 describes the LBP and LBP histogram feature, and different classifiers. The experiment setup and the result part are elaborately described in the section 3 and section 4 respectively. Section 5 gives the conclusion.

## 2. PRESENT WORK

### 2.1 Local Binary Patterns(LBP)

LBP is a very powerful feature for texture classification. It was first described by T. Ojala et al. [8][9]. Due to its gray scale and rotation invariance property, this feature has been used successfully in different domains of computer vision like face detection, facial expression recognition, brain MR image analysis etc.

This operator is invariant against any monotonic transformation of the gray scale of an image.LBP value of a particular pixel is always calculated by considering the pixel property of its neighbourhoods and this is the significance of using the term "Local". It is described later how this feature can be defined using '0' and '1' only and ultimately form a 0-1 pattern, that's why it is called "Binary Pattern".

LBP can be defined by using a texture T in a local neighbourhood of a pixel and the joint distribution of the gray levels of P neighbourhood image pixels. Here the parameter used in the pixel value may be its intensity value directly or it may be some other feature like gradient of that image pixel as described in the paper [10]. Normally all the description will be given here just by considering the gray level of intensity of the pixels.

So texture T can be defined in the following way

$$T = t(n_c, n_0, n_1, n_2 \ldots n_{P-1}); \quad (1)$$

Where $n_c$ denotes the gray level of the centered pixel of the local neighbourhood for which the LBP will be calculated. $n_p$ (i=0,1,2,…,P-1) corresponds to the gray values of equally spaced pixels on a circle of radius R (R>0) that form a circularly symmetric neighbour set. Here 'P' is called angular resolution and 'R' is called spatial resolution.
If the coordinate of $n_c$ is (0, 0), then the coordinate point of $n_i$ is given by

$(-R\sin(2\pi i/P), R\cos(2\pi i/P))$

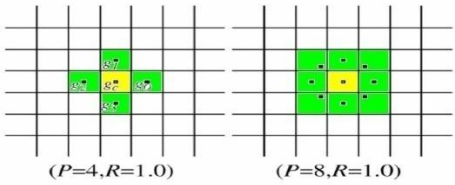

**Fig 1: Circularly symmetric neighbour sets (P: angular resolution, R: spatial resolution)**

Fig. 1 illustrates circularly symmetric neighbour sets for various (P, R). From the equation (1) gray scale invariance can be achieved by doing the following steps.

$T = t(n_c, n_0, n_1, n_2 \ldots n_{P-1});$
$T = t(n_c, n_0 - n_c, n_1 - n_c, n_2 - n_c \ldots n_{P-1} - n_c);$ step 1
$T \approx t(n_c) \, t(n_0 - n_c, n_1 - n_c, n_2 - n_c \ldots n_{P-1} - n_c);$ step 2
$T \approx t((n_0 - n_c), (n_1 - n_c), (n_2 - n_c) \ldots (n_{P-1} - n_c));$ step 3

In the first step the gray level of the centered pixel is subtracted from all the pixels of neighbourhood without losing its own gray level information. Since $(n_i - n_c)$ is independent of the value of $n_c$, we can write the second expression as the product of two terms as shown in the third expression above. We ignore the term $t(n_c)$, as shown in step3, since gray level changes between the central pixel and the local neighbourhood pixels are sufficient to represent a texture pattern.
Now if a constant region is found in any portion of an image, the differences are zero in all directions, whereas, for a spot, the differences are high in all directions [9].
Signed differences $(n_i - n_c)$ are not affected by changes in mean luminance because if a pixel is affected by noise then there is a high chance that its neighbourhood pixels will be affected by it more or less in same manner. Therefore subtracting this value indirectly removes the noise effect a little bit. Hence, the joint difference distribution is invariant against gray-scale shifts. We achieve invariance with respect to gray scale by considering just the signs (either '+' or '−'and '0') of the differences instead of their exact values: Hence it can be written as follows:

$T \approx t(\text{sign}(n_0 - n_c), \text{sign}(n_1 - n_c), \text{sign}(n_2 - n_c) \ldots \text{sign}(n_{P-1} - n_c));$

Where $\text{sign}(x) = 0$ if $x \leq 0$
$= 1$ if $x > 0$

So one binary pattern is calculated for a particular pixel and from this texture definition we can very easily calculate the numeric value (decimal) of LBP by the following formula:

$$LBP_{P,R} = \sum_{i=0}^{P-1} \cdot (\text{sign}(n_i - n_c) 2^i)$$

Rotation Invariance:

Normally there are $2^P$ values for a P-bit LBP pattern. For example if we use P=4 then there are 16 possible LBPs we can get ranging from 0000 to 1111.
If an image is rotated, then the neighbourhood pixels will be same but the relative positions will be different, hence we will get different LBP for the same pixel. So this operator is not rotation invariant.

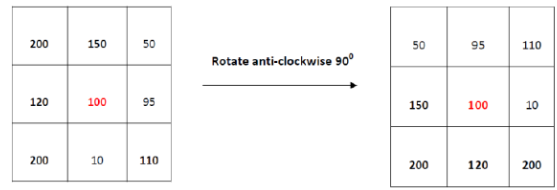

**(a) An image rotated anti clock wise by $q_0$**

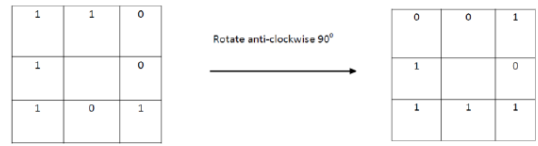

**b) Change in LBP due to this rotation**
**Fig 2: Example of rotation**

For example, from the Fig 2 we can clearly see that LBP value for the previous Pattern is → $(11001011)_2$ (starting from NW corner) and after rotating it $90^0$ anti-clockwise LBP value turns $(00101111)_2$
So to make it rotation invariant we will again define the LBP operator is defined in the following way.

$LBP^{ri}_{P,R} = \min\{ROR(LBP_{P,R}, i)\} \mid i = 0, 1, \ldots, P-1$

Where ROR(x,i) performs a circular bit-wise 'i' times right shift on the P bit number x. In terms of image pixels, this simply corresponds to rotating the neighbour set clockwise so as to get maximum number of zeros in the beginning of the binary string [9] 'min' operator will just take the minimum decimal values from different patterns. Instead of using min we may use 'max' (just opposite of the min) operator. Then also our goal of making it a rotation invariant will be achieved. This operator is also known as LBPROT as designed in the paper [7].

Uniform Pattern:

Sometimes this rotation invariant feature cannot give good discrimination [9]. Again among this rotation invariant patterns some patterns are dominant and are fundamental properties of texture. We call these fundamental patterns "uniform" as they have one thing in common, namely, uniform circular structure that contains very few spatial transitions.

A binary pattern is called "uniform" if it contains at most 2 spatial transitions (bitwise 0/1 changes). Based on this uniformity concept, a new LBP value ( $LBP^{riu}_{P,R}$ ) can be computed by summing up the bit values of a rotation invariant binary pattern if it is uniform, or a miscellaneous label P+1 can be assigned if it is non-uniform.

The uniform property is defined as follows:

$$LBP^{riu2}_{P,R} = \sum_{p=0}^{p-1} sign((n_p - n_c)) \text{ if } U(LBP_{P,R}) \leq 2$$

$$= (P+1)$$

Otherwise  Where

$$U(LBP_{P,R}) = |sign(n_{P-1} - n_c) - sign(n_0 - n_c)|$$
$$+ \sum_{p=1}^{p-1} |sign((n_p - n_c)) - sign((n_{p-1} - n_c))|$$

## 2.2 LBP Histogram

In statistical analysis, a histogram is a representation of the distribution of different parameters in an event. This idea has been extended to image analysis. For example, in case of digital images, a colour histogram represents the number of pixels corresponding to each colour / intensity value that spans the image's colour space.

Similarly we can store the LBP information of a particular image in histogram form and then analyze that histogram using some statistical operators for example Chi-square, Log statistic etc or we can apply some soft computing tools on it like SVM, MLP etc. If we just consider the normal LBP then it will have $2^P$ no of bins in the histogram. If LBP with uniform property is used then there will be P+2 bins (0 to P+1).

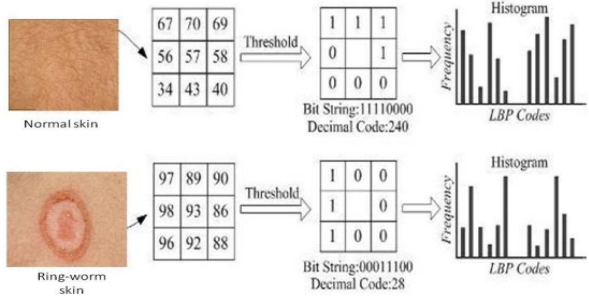

**Fig 3: Histogram formation from LBP pattern**

## 2.3 Different classifiers

### 2.3.1 Naive Bayes Classifier

The Naive Bayes Classifier technique[11] is based on the so-called Bayesian theorem and is particularly suited when the dimensionality of the inputs is high. Despite its simplicity, Naive Bayes can often outperform more sophisticated classification methods. Bayesian classification is an algorithm which allows us to categorize inputs probabilistically. Here in this experiment normal naive Bayesian classifier is used.

Abstractly, the probability model for this classifier is a conditional model

$$P(C|F_1, \ldots, F_n)$$

over a dependent class variable C with a small number of outcomes or classes. All the class elements have 'n' number of features namely $F_1, F_2, ..F_n$. If the value 'n' is not so large then this classifier provides good accuracy but on the contrary if it is so then the performance degrades. Even if the features take a large range of values then also its efficiency may reduce.

Using Bayes' theorem, we write

$$P(C|F_1, F_2, \ldots F_n) = \frac{P(C)p(F1,\ldots,Fn|C)}{p(F1,\ldots,Fn)}$$

In normal way the above equation can be written as:

Posterior = (prior X likelihood)/ (evidence)

In practice we are only interested in the numerator of that fraction, since the denominator does not depend on C and the values of the features $F_i$ are given, so that the denominator is effectively constant. The numerator is equivalent to the joint probability model

$$P(C, F_1, \ldots, F_n)$$

which can be rewritten as follows, using repeated applications of the definition of conditional probability[11].

$$p(C, F_1, \ldots, F_n)$$
$$= p(C) \; P(F_1, \ldots, F_n|C)$$
$$= p(C) \; p(F_1|C) \; p(F_2 \ldots F_n|C, F1)$$
$$= p(C) \; p(F_1|C) \; p(F_2|C, F1) \; p(F_3, F_4 \ldots F_n|C, F1, F2)$$
$$= p(C) \; p(F_1|C) \; p(F_2|C, F_1) \ldots p(Fn|C, F_1, F_2.F_{n-1})$$

Now Naïve conditional independence assumptions come into play. Assume that each feature $F_i$ is conditionally independent of every other feature $F_j$ for i≠j. This means that $p(F_i|C, F_j) = p(F_i|C)$; therefore the equation is

$$= p(C) \; p(F_1|C) \; p(F_2|C) \ldots p(F_n|C)$$

So ultimately we get

$$p(C, F_1, \ldots, F_n) = \prod_{i=1}^{n} p(C) \; p(F_i|C)$$
$$= p(C) \prod_{i=1}^{n} p(F_i|C)$$

For calculating the probability value of each numeric feature belongs to a particular class, the following probability density function is used.

$$f(x) = \frac{1}{\sqrt{2\pi}\sigma} e^{\frac{(x-\mu)^2}{2\sigma^2}}$$

Where 'μ', 'σ' are the mean and standard deviation of the training samples of a particular feature particular class respectively. 'x' is the feature value of the testing input-pattern.

### 2.3.2 Multi-Layer Perceptron (MLP)

An MLP is a feed-forward layered network of artificial neurons. Each artificial neuron in the MLP computes a special function on the weighted sum of all its inputs. The function may be a sigmoid or a polynomial or a hyperbolic tangent or a simple linear function. In our case we have used sigmoid function for each node in each layer. An MLP[12] consists of one input layer, one output layer and a number of hidden or intermediate layers. The output from every neuron in a layer of the MLP is connected to all inputs of each neuron in the immediate next layer of the same. The numbers of neurons in the input and the output layers of an MLP are chosen depending on the problem to be solved. For example in our experiment number of nodes in the input side is 160 that is equal to the number of features and in the output side it is 2 that is the number of classes. The number of neurons in hidden layer is determined by a trial and error method at the time of its training. An ANN requires training to learn an unknown input-output relationship to solve a problem.

The MLP classifier designed for the present work is trained with the Back Propagation (BP) algorithm. It minimizes the sum of the squared errors for the training samples by conducting a gradient descent search in the weight space. The number neurons in a hidden layer in the same are also adjusted during its training.

### 2.3.3 Support Vector Machine

Recently Support Vector Machine has been used successfully for pattern recognition and regression tasks formulized under the concept of structural risk minimization rule [13] It was mainly designed for binary classification, in order to construct an optimal hyper-plane, to maximize the margin of separation between the negative and positive data set. Although, SVM is used for two class pattern classification problem but multi-class problem can also be solved by extending the binary classification to multi class classification.

For the Support Vector Machine classifier, an open source software LibSVM tool is used. In general, a classification task usually involves with training and testing data which consist of some data instances. Each instance in the training set contains one "target value" (class labels) and several "attributes" (features). The goal of SVM is to produce a model which predicts target value of data instances in the testing set which are given only the attributes. Before considering the data directly from the linearly scaling each attribute to the range [-1, +1] or [0, 1].

Given a training set of instance-label pairs $(x_i, y_i)$;
$i =1,…., N$; where $x_i \in R_n$ and $y \in \{1, -1\}$, the support vector machines (SVM) require the solution of the following optimization problem:

min $(w, w_0, \epsilon,)$  $\frac{1}{2} w^T w + C \sum_{i=1}^{N} \epsilon_i$

Subject to  $y_i(w^T x_i + b) \geq 1 - \epsilon_i$,
$\epsilon_i \geq 0$

It's not difficult to generalize this linear program to the nonlinear case replacing xi with a nonlinear function M(xi):

min $(w, w_0, \epsilon,)$  $\frac{1}{2} w^T w + C \sum_{i=1}^{N} \epsilon_i$

Subject to  $y_i(w^T M(x_i) + b) \geq 1 - \epsilon_i$,
$\epsilon_i \geq 0$

Here expression for maximum margin is given as [13]

$$\text{margin} \equiv \arg\min_{\mathbf{x} \in D} d(\mathbf{x}) = \arg\min_{\mathbf{x} \in D} \frac{|\mathbf{x} \cdot \mathbf{w} + b|}{\sqrt{\sum_{i=1}^{d} w_i^2}}$$

The above illustration is the maximum linear classifier with the maximum range. Here training vectors xi are mapped into a higher dimensional space by the function M. Then SVM finds a linear separating hyper plane with the maximal margin in this higher dimensional space.

Furthermore, $k(x_i, x_j) \equiv M(x_i)^T M(x_j)$ is called the kernel function. In the current work, we have used the RBF kernel and polynomial kernel and the corresponding expressions are given below:

$$k(\mathbf{x_i}, \mathbf{x_j}) = \exp(-\gamma \|\mathbf{x_i} - \mathbf{x_j}\|^2)$$
$$k(\mathbf{x_i}, \mathbf{x_j}) = (\mathbf{x_i} \cdot \mathbf{x_j} + 1)^d$$

First one is for RBF and the next one is for polynomial.

## 3. EXPERIMENTAL SETUP

### 3.1 Preparation of database

For the experiment we don't have any direct database that's why different images of ringworm skin and normal skin have been collected from internet. These collected images are not directly used in the experiment. In order to prepare database of diseased skin and normal skin we have undertaken a sequence of steps. The processing steps of the images are described later in the same section. The database can be downloaded from the link www.cmaterju.org on request.

A total of 140 images are used for the experiment. 70 images are of ringworm affected skin and remaining 70 images are of normal skin. We have trained different classifiers using 50% of images from the ring worm positive and negative image sets. Thus our training set is formed with 70 images with equal number of images from each class.
The remaining images from the two sets are used as test data set for our experiment.

### 3.1.1. Preprocessing steps

In the processing steps we have done the following things.

i. Resizing all the images. All the images are of (144 x 144) pixels. For this we have used fotoxx software in Linux (Ubuntu) OS.
ii. Cropping the specified zones of the diseased skin.
iii. Converting those images into gray images. Here '.pgm' format is used for the actual experiments.

### 3.1.2. Decomposition of skin images

After the pre-processing, LBP histogram is calculated. For this, different region based approach is also considered. Each image is decomposed into 16 zones or regions R0, R1 ...R15. and each region is assigned equal weight because unlike face recognition, here all the regions are of equal importance. Here P is chosen as 8 and R as 1.

For each region 10 histogram features are generated. Thus total of 160 features (16x10) are calculated for each texture pattern.

So normalized feature vector is itself the region feature vector. By concatenating the entire region feature vector together, global information of the entire image can be obtained.

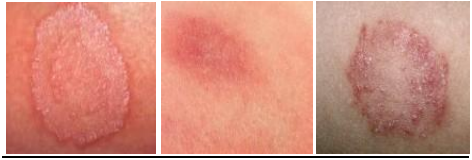

Fig 4: sample of Ringworm images

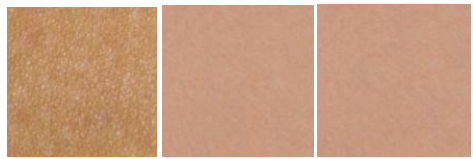

Fig 5: sample of normal skins

## 4. RESULT

For the present work, Multi Layer Perceptron(MLP) with one hidden layer is chosen. This is mainly to keep the computational requirement of the same, low without affecting its function approximation capability . To design the MLP for classification of ring worm disease, Back Propagation (BP) learning algorithm with learning rate ($\eta$) = 0.8 and momentum term ($\alpha$) = 0.7 is used here for training the classifier with varying number of neurons in its hidden layer. On the other hand, to implement SVM classifier we used an open source software LibSVM tool [14] .Among different existing kernels in LIBSVM we were used Radial basis Function (RBF) kernel with 0.5 gamma value.

In case of Bayesian classifier no parameters has to be set for the experiment externally. Just the mean and standard deviation of each individual feature for the training data has been calculated using the standard formula. After that for each testing pattern the probability for belonging to both classes was calculated and based on this value all the patterns are classified.

We have used 10 fold cross validation techniques to validate the database. The results are shown in Table 1. From the table we have found that MLP classifier gives maximum average success rate of 95.71% with 13.55 standard variations. Whereas, Bayesian classifier gives lowest average success rate of 70% with 17.10% standard deviation. SVM provides 74.28% success rate for these

We have applied the above methodologies on the designed dataset described in section 3. After training the classifiers we have used the test set to get actual recognition. Table no 2 shows the result of different classifiers on test data. From the Table no 2 we have found 72.85%, 94.28%, 90% success rate using Bayesian, MLP and SVM classifier respectively. And using majority voting scheme, we have obtained 91.42% success rate. The Fig 6 shows the number of classified data by different classifiers and their intersections and unions. Fig 7 shows some samples of misclassified data.

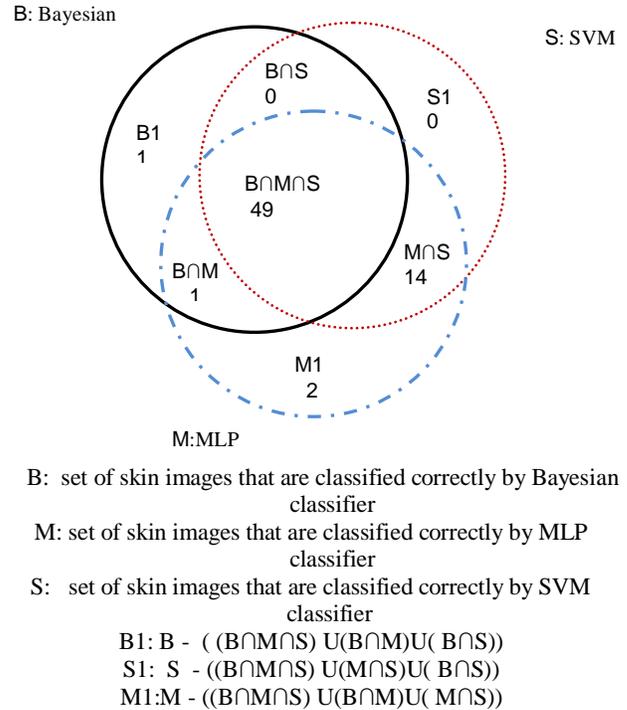

B: set of skin images that are classified correctly by Bayesian classifier
M: set of skin images that are classified correctly by MLP classifier
S: set of skin images that are classified correctly by SVM classifier
B1: B - ( (B∩M∩S) U(B∩M)U( B∩S))
S1: S - ((B∩M∩S) U(M∩S)U( B∩S))
M1:M - ((B∩M∩S) U(B∩M)U( M∩S))

Fig 6 Distribution of correctly and incorrectly classified samples by Bayesian, MLP and SVM

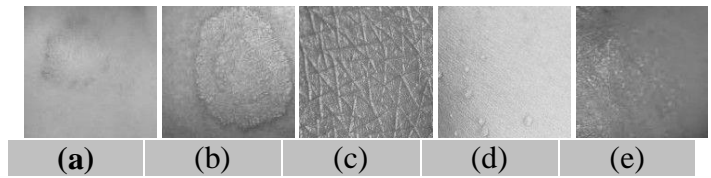

Fig 7: (a),(b)Ringworm infected skins classified as Ringworm negative;(c),(d),(e) Normal skin images classified as Ringworm positive

**Table 1 result of 10 fold cross validation**

| CV Fold / Classifier | Bayesian | MLP | SVM |
|---|---|---|---|
| Fold 1 | 71.43% | 57.14% | 57.14% |
| Fold 2 | 85.71% | 100.0% | 71.43% |
| Fold 3 | 85.71% | 100.0% | 100.0% |
| Fold 4 | 85.71% | 100.0% | 71.43% |
| Fold 5 | 71.43% | 100.0% | 100.0% |
| Fold 6 | 42.85% | 100.0% | 57.14% |
| Fold 7 | 42.85% | 100.0% | 71.43% |
| Fold 8 | 57.14% | 100.0% | 57.14% |
| Fold 9 | 71.43% | 100.0% | 71.43% |
| Fold 10 | 85.71% | 100.0% | 85.71% |
| Mean | 70.0% | 95.71% | 74.28% |
| Standard deviation | 17.10 | 13.55 | 16.21 |

**Table 2 result on test data using different classifiers and their majority voting**

|  | Bayesian classifier | MLP classifier | SVM classifier | Majority voting |
|---|---|---|---|---|
| Success rate | 72.85% | 94.28% | 90% | 91.42% |

## 5. CONCLUSION

As best of our knowledge there is no prior effort for automatic detection of ringworm. Hence, it is not possible to compare the results with others. On the other hand, LBP based features for medical images are a novel technique. It is low cost technique and does not require any special imaging devices. Some patterns shown in Fig 7 is quite difficult to recognize properly. Therefore the average success rate we have obtained here is quite satisfactory. The success rate may be improved with addition of some other features along with LBP. Use of other classifiers like RBF instead of Bayesian classifier may be ameliorated the success rate.

## ACKNOWLEDGEMENT

Authors are thankful to the "Center for Microprocessor Application for Training Education and Research", "Project on Storage Retrieval and Understanding of Video for Multimedia" of Computer Science & Engineering Department, Jadavpur University, for providing infrastructure facilities during progress of the work.